\def\year{2015}
\documentclass[letterpaper]{article}
\usepackage{aaai16}
\usepackage{times}
\usepackage{helvet}
\usepackage{courier}
\usepackage{amsmath}
\usepackage{amsfonts}
\usepackage{amssymb}
\usepackage{amsthm}
\usepackage{ dsfont }
\usepackage[table]{xcolor}
\usepackage{tikz}
\usepackage[ruled,vlined]{algorithm2e}

\newtheorem{ex}{Example}
\newtheorem{definition}{Definition}
\newtheorem{prop}{Proposition}

\newcommand \vrai{\checkmark}
\newcommand \faux{$\times$}
\newcommand \lex{lex}
\newcommand \iso{\gamma}
\newcommand{\propriete}[2]{\newline {\noindent \bf (#1)} \hfill #2 \hfill \null \\}
\newcommand{\proprieteligne}[2]{\newline {\noindent \bf (#1)}  #2 \\}
\newcommand{\attack}[2]{R^{-}_{#1}(#2)}
\newcommand{\defense}[2]{R^{+}_{#1}(#2)}
\newcommand{\battack}[2]{B\attack{#1}{#2}}
\newcommand{\bdefense}[2]{B\defense{#1}{#2}}
\newcommand\N{\mathds{N}}

\newcommand{\succcat}{\succ^{\mbox{\tiny Cat}}}
\newcommand{\succeqcat}{\succeq^{\mbox{\tiny Cat}}}
\newcommand{\succsaf}{\succ^{\mbox{\tiny SAF}}}
\newcommand{\succeqsaf}{\succeq^{\mbox{\tiny SAF}}}
\newcommand{\succdbs}{\succ^{\mbox{\tiny Dbs}}}
\newcommand{\succeqdbs}{\succeq^{\mbox{\tiny Dbs}}}
\newcommand{\succbbs}{\succ^{\mbox{\tiny Bbs}}}
\newcommand{\succeqbbs}{\succeq^{\mbox{\tiny Bbs}}}
\newcommand{\succtuple}{\succ^{\mbox{\tiny T}}}
\newcommand{\succeqtuple}{\succeq^{\mbox{\tiny T}}}
\newcommand{\prectuple}{\prec^{\mbox{\tiny T}}}
\newcommand{\preceqtuple}{\preceq^{\mbox{\tiny T}}}
\newcommand{\succmt}{\succ^{\mbox{\tiny M\&T}}}
\newcommand{\succeqmt}{\succeq^{\mbox{\tiny M\&T}}}
\newcommand{\simeqmt}{\simeq^{\mbox{\tiny M\&T}}}

\begin{document}
%
\title{A Comparative Study of Ranking-based Semantics for Abstract Argumentation}
\author{Elise Bonzon\\LIPADE\\Universit\'e Paris Descartes\\France\\bonzon@parisdescartes.fr\\
\And
J\'er\^ome Delobelle\\CRIL, CNRS\\Universit\'e d'Artois\\France\\delobelle@cril.fr\\
\And 
S\'ebastien Konieczny\\CRIL, CNRS\\Universit\'e d'Artois\\France\\konieczny@cril.fr\\
\And
Nicolas Maudet\\
Sorbonne Universit\'es \\
UPMC Univ Paris 06, CNRS \\
LIP6, UMR 7606 \\
75005 Paris\\
nicolas.maudet@lip6.fr\\
}
\maketitle
\begin{abstract}
\begin{quote}
Argumentation is a process of evaluating and comparing a set of arguments. 
A way to compare them consists in using a ranking-based semantics which rank-order arguments from the most to the least acceptable ones.
Recently, a number of such semantics have been proposed independently, often associated with some desirable properties.
However, there is no comparative study which takes a broader perspective.
This is what we propose in this work.
We provide a general comparison of all these semantics with respect to the proposed properties.
That allows to underline the  differences of behavior between the existing semantics.
\end{quote}
\end{abstract}

\section{Introduction}
Argumentation consists in reasoning with conflicting information based on the exchange and evaluation of interacting arguments.
The most popular way to represent argumentation process was proposed by Dung \shortcite{dung95} with argumentation frameworks modelized by binary graphs, where the nodes represent the arguments, and the edges represent the attacks between them.
From these argumentation frameworks, several semantics indicating which sets of arguments, called extensions, are mutually compatible were proposed (see \cite{Baroni11} for an overview).
However, for applications with a big number of arguments, it can be problematic to have only two levels of evaluations (arguments are either accepted or rejected).  For instance, such a limitation can be questionable when using argumentation for debate platforms on the web (see \cite{LM11} for such a discussion).

In order to fix these problems, a solution consists in using semantics that distinguish arguments not with the classical accepted/rejected evaluations, but with a large number of levels of acceptability. A lot of these semantics, called ranking-based semantics, were proposed in recent years \cite{AB13,CLS05,LM11,MT08,G12} with, for each semantics, different behaviour and logical properties. However, all these semantics have never been compared between them.

This is what we propose in this work. We study the existing ranking-based semantics in the literature (focusing on the semantics that return a unique ranking between arguments) in the light of the proposed properties. That allows us to underline the differences of behavior between those semantics, and to propose a better reading of the different choices one has on this matter.

The paper is organized as follows. The following section gives the relevant background regarding abstract argumentation and ranking-based semantics. The next section presents the different properties that have been introduced in the literature, whereas the next one formally introduces the existing ranking-based semantics. 
Note that due to space constraints, we can not recall all the details and justifications of semantics and properties of the literature, but the reader can find them in the corresponding papers. In the next section we discuss the different properties and compare the semantics, and the final section concludes.

\section{Preliminaries}
In this section, we start by briefly recalling what is a Dung's abstract argumentation framework \cite{dung95}.
\begin{definition}
An \emph{\textbf{argumentation framework (AF)}} is a pair $F = \langle A,R \rangle$ with $A$ a set of arguments and $R$ a binary relation on $A$, i.e.\ $R \subseteq A \times A$, called the \emph{\textbf{attack relation}}.
A set of arguments $S \subseteq A$ attacks an argument $b$ $\in$ $A$, if there exists $a \in S$, such that $(a,b) \in R$. We note $Arg(F)=A$.
\end{definition}
Let $\mathds{AF}$ be the set of all argumentation frameworks. For two AFs $F = \langle A,R \rangle$ and $G = \langle A',R' \rangle$, we define the union $F \cup G = \langle A \cup A', R \cup R' \rangle$.

We can now introduce some useful notions in order to formalize properties of argumentation frameworks.

\begin{definition}
Let $F = \langle A,R \rangle$ be an AF 
and $a,b \in A$. 
A {\bf path} $P$ from $b$ to $a$, noted $P(b,a)$, is a sequence $s=\langle a_{0},\dots,a_{n} \rangle$ of arguments such that $a_{0} = a$, $a_{n} = b$ and $\forall i < n, (a_{i+1},a_{i}) \in R$. 
We denote by $l_{P} = n$ the {\bf length} of P. 
A {\bf defender} (resp. {\bf attacker}) of $a$ is an argument situated at the beginning of an even-length (resp. odd-length) path. 
We denote the multiset of defenders and attackers of $a$ by $\defense{n}{a} = \{ b\ |\ \exists P(b,a)\ with\ l_{P} \in 2\mathds{N}\}$ and $\attack{n}{a} = \{ b\ |\ \exists P(b,a)\ with\ l_{P} \in 2\mathds{N}+1\}$ respectively.  The  {\bf direct attackers} of $a$ are arguments in $\attack{1}{a}$.  An argument $a$ is {\bf defended} if $\defense{2}{a} \not= \emptyset$.\\
A {\bf defense root} (resp. {\bf attack root}) is a non-attacked defender (resp. attacker). 
We denote the multiset of defense roots and attack roots of $a$ by $\bdefense{n}{a} = \{ b \in \defense{n}{a} \ |\ |\attack{1}{b}|=0 \}$ and $\battack{n}{a} = \{ b \in \attack{n}{a}\ |\ |\attack{1}{b}|=0\}$ respectively.
A path from $b$ to $a$ is a {\bf defense branch} (resp. {\bf attack branch}) if $b$ is a defense (resp. attack) root of $a$.
Let us note $\bdefense{}{a}=\bigcup_{n} \bdefense{n}{a}$ and $\battack{}{a}=\bigcup_{n} \battack{n}{a}$.
\end{definition}

The {\it connected components} of an AF are the set of largest subgraphs of AF, denoted by $cc(AF)$, where two arguments are in the same component of AF if and only if there is some path (ignoring the direction of the edges) between them.

In Dung's framework \cite{dung95}, the {\it acceptability} of an argument depends on its membership to some sets, called extensions. Another way to select a set of acceptable arguments is to {\it rank} arguments from the most to the least acceptable ones. 
Ranking-based semantics aim at determining such a ranking between arguments. 
\begin{definition}
	A {\bf ranking-based semantics} $\sigma$ associates to any argumentation framework AF = $\langle A,R \rangle$  a ranking $\succeq^{\sigma}_{AF}$ on A, where $\succeq^{\sigma}_{AF}$ is a preorder (a reflexive and transitive relation) on $A$. $a \succeq^{\sigma}_{AF} b$ means that $a$ is at least as acceptable as $b$ ($a \simeq_{AF}^{\sigma} b$ is a shortcut for $a \succeq^{\sigma}_{AF} b$ and $b \succeq^{\sigma}_{AF} a$, and
$a \succ^{\sigma}_{AF} b$ is a shortcut for $a \succeq^{\sigma}_{AF} b$ and $b \nsucceq^{\sigma}_{AF} a$).
\end{definition}
When there is no ambiguity about the argumentation framework in question, we will use $\succeq^{\sigma}$ instead of $\succeq^{\sigma}_{AF}$.

Finally, we need to introduce the notion of lexicographical order in order to define some ranking-based semantics.
\begin{definition}
	A {\bf lexicographical order} between two vectors of real number $V = \langle V_{1}, \dots, V_n \rangle$ and $V' = \langle V'_{1}, \dots, V'_n \rangle$, is defined as $V \succeq_{lex} V'$ iff $\exists i \leq n$ s.t. $V_{i} \ge V'_{i}$ and $\forall j<i, V_{j} =  V'_{j}$.
	
\end{definition}

\section{Properties}
Let us recall the logical properties proposed in the literature for ranking-based semantics.
Please note that all the properties are not mandatory (we will see later that some of them are incompatible), but we want to give all of them for completeness and since we will check them for the existing ranking-based semantics.
Unless stated explicitly, all the properties are defined for a ranking-based semantics $\sigma$, $\forall AF \in \mathds{AF}$ and $\forall a,b \in Arg(AF)$.

\begin{definition}
An {\bf isomorphism} $\iso$ between two argumentation frameworks AF = $\langle A,R \rangle$ and AF' = $\langle A',R' \rangle$ is a bijective function $\iso : A \rightarrow A'$ such that $\forall x, y \in A$, $(x,y) \in R$ iff $(\iso(x),\iso(y)) \in R' $.  With a slight abuse of notation, we will note $AF' = \iso(AF)$. 
\end{definition}

\noindent \textbf{Abstraction.} The ranking on $A$ should be defined only on the basis of the attacks between arguments.
\proprieteligne{Abs}{%
Let $AF, AF' \in \mathds{AF}$. For any isomorphism $\iso$ s.t. $AF' = \iso(AF)$, we have $a \succeq^{\sigma}_{AF} b\ \text{iff}\ \iso(a) \succeq^{\sigma}_{AF'} \iso(b)$}

\noindent \textbf{Independence.}
The ranking between two arguments $a$ and $b$ should be independent of any argument that is neither connected to $a$ nor to $b$.
\proprieteligne{In}{$\forall AF' \in cc(AF)$, $\forall a,b \in Arg(AF')$, \\ \hspace*{3cm}   $a \succeq^{\sigma}_{AF'} b$ $\Rightarrow$ $a \succeq^{\sigma}_{AF} b$}

\noindent We may have expectations regarding the best and worst arguments that we may find in an AF:

\noindent \textbf{Void Precedence.} A non-attacked argument is ranked strictly higher than any attacked argument.
\propriete{VP}{$\attack{1}{a} = \emptyset$ and $\attack{1}{b} \ne \emptyset \Rightarrow\ a \succ^{\sigma} b$}

\noindent \textbf{Self-Contradiction.} 
A self-attacking argument is ranked lower than any non self-attacking argument.
\propriete{SC}{$(a,a) \notin R$ and $(b,b) \in R\ \Rightarrow\ a \succ^{\sigma} b$}

\noindent The following local properties are concerned with the direct attackers, or defenders, of arguments: 		

\noindent \textbf{Cardinality Precedence.} %
The greater  the number of direct attackers for an argument, the weaker the level of acceptability of this argument.
\propriete{CP}{$|\attack{1}{a}| < |\attack{1}{b}| \Rightarrow\ a \succ^{\sigma} b$}
	
\noindent \textbf{Quality Precedence.} %
The greater  the acceptability of one direct attacker for an argument, the weaker  the level of acceptability of this argument.
\proprieteligne{QP}{$\exists c \in \attack{1}{b}$ s.t. $\forall d \in \attack{1}{a},\ c \succ^{\sigma} d \Rightarrow\ a \succ^{\sigma} b$}

Before defining the next properties, we need to introduce a relation that compares sets of arguments on the basis of their rankings \cite{AB13}:
\begin{definition}
	Let $\ge_{S}$ be a ranking on a set of arguments A. For any $S_{1},S_{2} \subseteq A$, $S_{1} \ge_{S} S_{2}$ is a {\bf group comparison} iff there exists an injective mapping f from $S_{2}$ to $S_{1}$ such that $\forall a \in S_{2}, f(a) \succeq a$. And $S_{1} >_{S} S_{2}$ is a strict group comparison iff $S_{1} \ge_{S} S_{2}$ and $(|S_{2}| < |S_{1}|$ or $\exists a \in S_{2}, f(a) \succ a)$. 
\end{definition}

\noindent \textbf{Counter-Transitivity.} If the direct attackers of $b$ are at least as numerous and acceptable as those of $a$, then $a$ is at least as acceptable as $b$. 
\propriete{CT}{$\attack{1}{b} \ge_{S} \attack{1}{a} \Rightarrow a \succeq^{\sigma} b$}
	
\noindent \textbf{Strict Counter-Transitivity.} If CT is satisfied and either the direct attackers of $b$ are strictly more numerous or acceptable than those of $a$, then $a$ is strictly more acceptable than $b$. 
\propriete{SCT}{$\attack{1}{b} >_{S} \attack{1}{a} \Rightarrow a \succ^{\sigma} b$}

\noindent \textbf{Defense Precedence.} For two arguments with the same number of direct attackers, a defended argument is ranked higher than a non-defended argument.
\proprieteligne{DP}{$|\attack{1}{a}| = |\attack{1}{b}|, \defense{2}{a} \ne \emptyset$ and $\defense{2}{b} = \emptyset$ \\ \null \hfill $\Rightarrow\ a \succ^{\sigma} b$}

\begin{definition}
	Let $AF = \langle A,R \rangle$ and $a \in A$. The {\bf defense} of $a$ is {\bf simple} iff every defender of $a$ attacks exactly one direct attacker of $a$. The defense of $a$ is {\bf distributed} iff every direct attacker of $a$ is attacked by at most one argument.
\end{definition}

\noindent \textbf{Distributed-Defense Precedence.} 
The best defense is when each defender attacks a distinct attacker. 
\proprieteligne{DDP}{$|\attack{1}{a}| = |\attack{1}{b}|$ and $|\defense{2}{a}| = |\defense{2}{b}|$, if the defense of $a$ is simple and distributed and the defense of $b$ is simple but not distributed, then $a \succ^{\sigma} b$}

The following properties check if some change in an AF can improve or degrade the ranking of one argument. These properties have been proposed informally by Cayrol and Lagasquie-Schiex \shortcite{CLS05}, in the context of their semantics.  We propose a formalization that generalize them for any argumentation frameworks. We first define the addition of a defense/attack branch to an argument. 
\begin{definition}
	Let $AF = \langle A,R \rangle$, $a \in A$. 
	The {\bf defense branch added to $a$} is $P_{+}(a) = \langle A', R' \rangle$, with $A'=\{x_{0}, \dots, x_{n}\}$,  $n \in 2\mathds{N}$, $x_{0}=a$, $A \cap A' = \{a\}$,  and $R' =  \{(x_{i},x_{i-1}) \mid i \le n\}$. 
	The {\bf attack branch added to $a$}, denoted $P_{-}(a)$ is defined similarly except that the sequence is of odd length (i.e. $n \in 2\mathds{N}+1$).

\end{definition}

The following properties are defined $\forall AF, AF^{\gamma} \in \mathds{AF}$ such that exists an isomorphism $\iso$ with $AF^{\gamma} = \iso(AF)$, and $\forall a \in Arg(AF)$.
We use $AF^{\gamma}$ as a clone of $AF$.

\noindent{\textbf{Strict addition of Defense Branch.}}
Adding a defense branch to any argument improves its ranking.
\proprieteligne{$\oplus$DB}{If $AF^{\star}=AF \cup AF^{\gamma} \cup P_{+}(\iso(a))$, then $\iso(a) \succ_{AF^{\star}}^{\sigma} a$}

\noindent{\textbf{Addition of Defense Branch.}}
It could make sense to treat differently non-attacked arguments. So in \cite{CLS05}, this property is defined in a more specific way: adding a defense branch to any attacked argument improves its ranking.
\proprieteligne{+DB}{If $AF^{\star}=AF \cup AF^{\gamma} \cup P_{+}(\iso(a))$ and $|R^-_{1}{(a)}| \neq 0$, then $\iso(a) \succ_{AF^{\star}}^{\sigma} a$}

\noindent{\textbf{Increase of Attack branch.}}
Increasing the length of an attack branch of an argument improves its ranking.
\proprieteligne{$\uparrow$AB}{If $b \in \battack{}{a}$, $b \notin \bdefense{}{a}$ and $AF^{\star}=AF \cup AF^{\gamma} \cup P_{+}(\iso(b))$, then $\iso(a) \succ_{AF^{\star}}^{\sigma} a$}

\noindent{\textbf{Addition of Attack Branch.}}
Adding an attack branch to any argument degrades its ranking.
\propriete{+AB}{If $AF^{\star}=AF \cup AF^{\gamma} \cup P_{-}(\iso(a))$, then $a \succ_{AF^{\star}}^{\sigma} \iso(a)$}

\noindent{\textbf{Increase of Defense branch.}}
Increasing the length of a defense branch of an argument degrades its ranking.
\proprieteligne{$\uparrow$DB}{If $b \in BR^{+}(a)$, $b \notin \battack{}{a}$ and $AF^{\star}=AF \cup AF^{\gamma} \cup P_{+}(\iso(b))$, then $a \succ_{AF^{\star}}^{\sigma} \iso(a)$}

One can find the properties Abs, In, VP, DP, CT, SCT, CP, QP and DDP in \cite{AB13}, the properties In, VP and SC in \cite{MT08} and the property VP in \cite{CLS05}.

To this set of properties from the literature we want to add some other important properties. 
\\
\noindent \textbf{Total.} All pairs of arguments can be compared.
\propriete{Tot}{$a \succeq^{\sigma} b\ or\ b \succeq^{\sigma} a$}

The next property states that all the non-attacked arguments should have the same ranking. 

\noindent\textbf{Non-attacked Equivalence.} All the non-attacked argument have the same rank.
\propriete{NaE}{$\attack{1}{a} = \emptyset\ and\ \attack{1}{b} = \emptyset \Rightarrow\ a \simeq^{\sigma} b$}
	
The last property describes the behavior adopted by a semantics concerning the notion of defense, and can be viewed as some kind of compatibility with usual Dung's semantics. 
The idea is that a defended argument is always better than an attacked argument.
\\
\noindent \textbf{Attack vs Full Defense.} An argument without any attack branch is ranked higher than an argument only attacked by one non-attacked argument.
\proprieteligne{AvsFD}{$AF \mbox{ is acyclic}, |\battack{}{a}|=0, |\attack{1}{b}| = 1$ and $|\defense{2}{b}| = 0 \Rightarrow a \succ^{\sigma} b$}

Let us now check the incompatibilities/dependencies between properties.
\begin{prop}
\label{prop:incomp}
	For every ranking-based semantics, the following pairs of properties are not compatible :
	\begin{itemize}
		\item  {CP} and  {QP} \cite{AB13}
		\item  {CP} and  {AvsFD}
		\item {CP} and  {+DB}
		\item {VP} and {$\oplus$DB}
	\end{itemize}
	The following properties are not independent :
	\begin{itemize}
		\item {SCT} implies {VP} \cite{AB13}
		\item {CT} and {SCT} imply {DP} \cite{AB13}
		\item {SCT} implies {CT}
		\item {CT} implies {NaE}
		\item {$\oplus$DB} implies {+DB}
	\end{itemize}
\end{prop}

\section{Existing Ranking-based Semantics}

\subsection{Categoriser}
Besnard and Hunter \shortcite{BH01} propose a {\it categoriser} function which assigns a value to each argument, given the value of its direct attackers.
\begin{definition}[Besnard and Hunter 2001]
Let $F = \langle A,R \rangle$ be an AF. 
The {\bf categoriser function} $Cat : A \rightarrow ]0,1]$ is defined as:
		\[
			Cat(a) = 
			\left\{
			\begin{array}{l l}
			1 & \ \mbox{if } \attack{1}{a} = \emptyset\\
			{\frac{1}{1+\sum_{c \in \attack{1}{a}}Cat(c)}} & \mbox{otherwise}\\
			\end{array}
			\right.
		\]
\end{definition}

\begin{definition}
The {\bf ranking-based semantics Categoriser} associates to any AF = $\langle A,R \rangle$  a ranking $\succeqcat_{AF}$ on $A$ such that $\forall a,b \in A, a \succeqcat_{AF} b$ iff $Cat(a) \ge Cat(b)$.
\end{definition}

\begin{ex}
Let $F=\langle A,R \rangle$ with $A = \{a,b,c,d,e\}$ and $R = \{(a,e),(b,a),(b,c),(c,e),(d,a),(e,d)\}$.

	\begin{figure}[!h]
	\begin{center}
		\begin{tikzpicture}[->,>=stealth,shorten >=1pt,auto,node distance=1cm,
 				 thick,main node/.style={circle,draw}]

 			 \node[main node] (1) {d};
 			 \node[main node] (2) [right of=1] {a};
  			 \node[main node] (3) [right of=2] {b};
  			 \node[main node] (4) [below of=1] {e};
  			 \node[main node] (5) [below of=3] {c};

   			 \draw[->,>=latex] (1) to (2);
    			 \draw[->,>=latex] (2) to (4);
  			 \draw[->,>=latex] (4) to (1);
  			 \draw[->,>=latex] (3) to (2);
  			 \draw[->,>=latex] (3) to (5);
  			 \draw[->,>=latex] (5) to (4);
    
		\end{tikzpicture} 
		\end{center}
		
		\caption{\label{figurePrems}An argumentation framework}
		\end{figure}
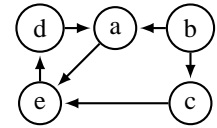
\label{ExPrems}
The categoriser of each argument are
$Cat(a)\approx0.38$ , $Cat(b) = 1$ , $Cat(c) = 0.5 $, $Cat(d) \approx 0.65$  and $Cat(e) \approx 0.53$ . So we obtain the ranking : $b \succcat d \succcat e \succcat c \succcat a$.
\end{ex}

This semantics takes into account only the value of the direct attackers to compute the strength of an argument. This is why the argument $e$ of the example, which is attacked twice but by arguments that are attacked by a non-attacked argument, is ranked higher than the argument $c$, which is attacked just once, but by a stronger argument.

\begin{prop}
	The ranking-based semantics Categoriser satisfies\footnote{The properties Abs, In, VP, DP, CT, SCT have already been checked by Pu et al. \shortcite{PLZL14}.} Abs, In, VP, DP, CT, SCT, $\uparrow$AB, $\uparrow$DB, +AB, Tot and NaE. The other properties are not satisfied.
\end{prop}

\subsection{Social Abstract Argumentation Framework}

Leite and Martins \shortcite{LM11} introduce an extension of Dung's abstract argumentation frameworks that include social voting on the arguments: the Social Abstract Argumentation Frameworks (SAF). They also propose a family of semantics where a model is a solution to the equation system\footnote{An equational approach was also proposed by Gabbay \shortcite{G12}. This method returns multiple solutions, and thus several rankings for one AF. This is why we do not consider this method in this paper.} with one equation for each argument, based on its social support and its direct attackers. In order to compare SAFs with the existing ranking-based semantics, we chose to ignore the social support of arguments by giving them the same value.

\begin{definition}
Let $F = \langle A,R \rangle$ be an AF and $S = \langle L, \tau,\curlywedge,\curlyvee, \neg \rangle$ be a (well-behaved) SAF semantic. 
The total mapping $M_{S}: A \rightarrow L$ is a {\bf social model} of $F$ under semantics $S$ such that $\forall a \in A$:
		$$M_{S}(a) = \tau(a) \curlywedge \neg \curlyvee\{M(a_{i}) : a_{i} \in \attack{1}{a}\}, \text{ where}$$
		\begin{itemize}
			\item $L$ is a totally ordered set with top $\top$ and bottom $\bot$ elements, containing all possible valuations of an argument;
			\item $\tau : A \rightarrow L$ is an attenuation factor. $\tau$ is monotonic w.r.t. the first argument and antimonotonic w.r.t the second argument;
			\item $\curlywedge : L \times L \rightarrow L$ combines the initial score with the score of direct attackers. $\curlywedge$ is continuous, commutative, associative, monotonic w.r.t. both arguments and $\top$ is its identity element;
			\item $\curlyvee : L \times L \rightarrow L$ aggregates the score of direct attackers. $\curlyvee$ is continuous, commutative, associative, monotonic w.r.t. both arguments and $\bot$ is its identity element;
			\item $\neg : L \rightarrow L$ restricts the value of the attacked argument. $\neg$ is antimonotonic, continuous, $\neg \bot = \top$, $\neg \top = \bot$ and $\neg \neg a = a$.
		\end{itemize}
\end{definition}

One possible  (well-behaved) SAF semantic 
proposed in \cite{LM11} is the {\it simple product semantic} 
$\emph{SP}_{\epsilon} = \langle [0,1], \tau_{\epsilon},\curlywedge,\curlyvee, \neg \rangle$ where $\tau_{\epsilon} = \frac{1}{1+\epsilon}$ (with $\epsilon > 0$, to ensure the uniqueness of the semantics), $x_{1} \curlywedge x_{2} = x_{1} \times x_{2}$ (Product T-Norm),  $x_{1} \curlyvee x_{2} = x_{1} + x_{2} - x_{1} \times x_{2}$ (Probabilistic Sum T-CoNorm) and $\neg x_{1} = 1 - x_{1}$.

	\begin{definition}
		The {\bf ranking-based semantics SAF} associates to any $AF = \langle A,R \rangle$  a ranking $\succeqsaf_{AF}$ on $A$ such that $\forall a,b \in A, a \succeqsaf_{AF} b$ iff $M_{S}(a) \geq M_{S}(b)$.
	\end{definition}

	\addtocounter{ex}{-1}
	\begin{ex}[cont.]
		With $\epsilon = 0.1$, we obtain $M_{\emph{SP}_{\epsilon}}(a)\approx0.07$ , $M_{\emph{SP}_{\epsilon}}(b)\approx0.91$ , $M_{\emph{SP}_{\epsilon}}(c)\approx0.08 $, $M_{\emph{SP}_{\epsilon}}(d)\approx0.20$  and $M_{\emph{SP}_{\epsilon}}(e)\approx0.78$. We obtain the ranking: $b \succsaf e \succsaf d \succsaf c \succsaf a$.
	\end{ex}
	
	As for the Categoriser semantics, the strength of attackers is more important than their numbers, and thus $e$ is preferred to $c$. However the impact of a defense branch on an argument is weaker with SAF than with Categoriser.
	
	\begin{prop}
	SAF satisfies Abs, In, VP, DP, CT, SCT, $\uparrow$AB, $\uparrow$DB, +AB, Tot and NaE. Other properties are not satisfied.
	\end{prop}

\subsection{Discussion-based semantics}
The Discussion-based semantics \cite{AB13} compares arguments by counting the number of paths ending to them. 
If some arguments are equivalent (they have the same number of direct attackers), the size of paths is recursively increased until a difference is found.
\begin{definition}
	Let $F = \langle A,R \rangle$ be an AF, $a \in A$, and $i \in \N$.
	\[
			Dis_{i}(a) = 
			\left\{
			\begin{array}{c l}
			-|\defense{i}{a}| & \mbox{if $i$ is odd}\\
			|\attack{i}{a}| & \mbox{if $i$ is even}\\
			\end{array}
			\right.
		\]
	The {\bf discussion count of $a$} is denoted $Dis(a) = \langle Dis_{1}(a),Dis_{2}(a),\dots \rangle$. 
	\end{definition}
	
	\begin{definition}
	The {\bf ranking-based semantics Dbs} associates to any 
	AF = $\langle A,R \rangle$  a ranking $\succeqdbs_{AF}$ on $A$ such that $\forall a,b \in A$, $a \succeqdbs_{AF} b$ iff $Dis(b) \succeq_{lex} Dis(a)$.
	
	\end{definition}
		
	\addtocounter{ex}{-1}
	\begin{ex}[cont.]
	\ 
		\begin{center}
			\begin{tabular}{|c|c|c|c|c|c|}
				\hline
				step & a & b & c & d & e \\
				\hline
				1 & 2 & 0 & 1 & 1 & 2\\
				\hline
				2 & -1 & 0 & 0 & -2 & -3\\
				\hline
			\end{tabular} 
		\end{center}
		Using the lexicographical order, one obtains the following ranking: $b \succdbs d \succdbs c \succdbs e \succdbs a$
		
	\end{ex}
	
	The number of attackers is more important than their strength, thus $c$ is here stronger than $e$.
	\begin{prop}
	Dbs satisfies Abs, In, VP, DP, CT, SCT, CP, $\uparrow$AB, $\uparrow$DB, +AB, Tot and NaE. The other properties are not satisfied.
	\end{prop}

\subsection{Burden-based semantics}
	The Burden-based semantics \cite{AB13} assigns, at each step $i$, a Burden number to every argument, that depends on the Burden numbers of its direct attackers. 
	
	\begin{definition}
		Let $F = \langle A,R \rangle$ be an AF, $a \in A$ and $i \in \N$. 
		\[
			Bur_{i}(a) = 
			\left\{
			\begin{array}{l l}
			1 & \mbox{if } i = 0\\
			1 + \sum_{b \in \attack{1}{a}}\frac{1}{Bur_{i-1}(b)} & \mbox{otherwise}\\
			\end{array}
			\right.
		\]
	The {\bf Burden number of $a$} is denoted $Bur(a) = \langle Bur_{0}(a), Bur_{1}(a),\dots \rangle$.
	\end{definition}
	
Two arguments are lexicographically compared on the basis of their Burden numbers.
	
	\begin{definition}
		The {\bf ranking-based semantics Bbs} associates to any $AF = \langle A,R \rangle$  a ranking $\succeqbbs_{AF}$ on $A$ such that $\forall a,b \in A$, $a \succeqbbs_{AF} b$ iff $Bur(b) \succeq_{lex} Bur(a)$.
	\end{definition}
	
	\addtocounter{ex}{-1}
	\begin{ex}[cont.] 
	\ 
		\begin{center}
			\begin{tabular}{|c|c|c|c|c|c|}
				\hline
				step & a & b & c & d & e \\
				\hline
				1 & 3 & 1 & 2 & 2 & 3\\
				\hline
				2 & 2.5 & 1 & 2 & 1.33 & 1.83\\
				\hline
			\end{tabular} 
		\end{center}
		Using the lexicographical order, one obtains the following ranking: $b \succbbs d \succbbs c \succbbs e \succbbs a$
		
	\end{ex}
	
	As on this example, Dbs and Bbs often return the same result. The main difference between these semantics is that Bbs satisfies DDP, so examples related to that kind of structures lead to distinct results.
		
	\begin{prop}
	Bbs satisfies Abs, In, VP, DP, CT, SCT, CP, DDP, $\uparrow$AB, $\uparrow$DB, +AB, Tot and NaE. The other properties are not satisfied.
	\end{prop}

\subsection{Valuation with tuples}
The semantics proposed by Cayrol and Lagasquie-Schiex \shortcite{CLS05} takes into account all the ancestors branches of an argument (defender and attacker) stored in tupled values :

\begin{definition}
		Let  $F = \langle A,R \rangle$ be an AF and $a \in A$. 
		Let $v_{p}(a)$ be the (ordered) tuple of even integers representing the lengths of all the defense branches of $a$, i.e.
		 $v_{p}(a)$ is the smallest ordered tuple such that $|\bdefense{n}{a}| = x \Rightarrow n \in_{x} v_{p}(a)$, where $\in_{x}$ means "appears at least $x$ times".
		Similarly let $v_{i}(a)$ be the (ordered) tuple of odd integers representing the lengths of all the attack branches of $a$, i.e. 
		 $v_{i}(a)$ is the smallest ordered tuple such that $|\battack{n}{a}| = x \Rightarrow n \in_{x} v_{i}(a)$.
		A {\bf tupled value} for $a$ is the pair  $v(a) = [v_{p}(a),v_{i}(a)]$.	
		\end{definition}

When cycles exist in the AF, some tuples can be infinite.
To calculate them, this method requires a highly involved process, that turn cyclical graphs into infinite acyclic graphs.
We thus consider this approach for acyclic graphs only, and denote it by $Tuples^{*}$.

Once the tupled value of each argument has been computed, one can compare them. To do so one has to compare the length of attack/defense branches and, in case of a tie, to compare the values inside each tuples (see Algorithm 1).

\setlength{\algomargin}{0.35em}
\begin{algorithm}
\caption{$Tuples^{*}$}
\SetKwInput{KwNotation}{Notation}
\SetAlgoVlined
\LinesNumbered
\KwIn{$v(a),w(b)$ two tupled values of arguments $a$ and $b$ }
\KwOut{A ranking $\succeqtuple$ between $a$ and $b$}
\BlankLine
\small{
\Begin{
	\lIf{v = w}{$a \succeqtuple b$ and  $b \succeqtuple a$}
	\Else{
		\If{$|v_{i}| = |w_{i}|$ {\bf and} $|v_{p}| = |w_{p}|$}
		{
			\lIf{$v_{p} \preceq_{\lex} w_{p}$ {\bf and} $v_{i} \succeq_{\lex} w_{i}$}{$a \succtuple b$}
			\Else{
				\mbox{\lIf{$v_{p} \succeq_{\lex} w_{p}$ {\bf and} $v_{i} \preceq_{\lex} w_{i}$}{$a \prectuple b$}}
				\lElse{$a \not\succeqtuple b$ and $a \not\preceqtuple b$}
			}
		}
		\Else{
			\lIf{$|v_{i}| \ge |w_{i}|$ {\bf and} $|v_{p}| \le |w_{p}|$}{$a \prectuple b$}
			\Else{
				\mbox{\lIf{$|v_{i}| \le |w_{i}|$ {\bf and} $|v_{p}| \ge |w_{p}|$}{$a \succtuple b$}}
				\lElse{$a \not\succeqtuple b$ and $a \not\preceqtuple b$}
			}
		}
	}

}}
\end{algorithm}
Let us remark that two arguments can be incomparable. It is the case, for example, if an argument has strictly more attack branches and more defense branches than another one. Consequently, this semantics returns a partial ranking between arguments.

As example 1 contains a cycle, we can not compute Tuples$^{*}$ on this running example.

	\begin{prop}
	The ranking-based semantics Tuples$^{*}$ satisfies Abs, In, VP, +DB,  $\uparrow$AB,  $\uparrow$DB, +AB, NaE and AvsFD. The other properties are not satisfied.
	\end{prop}

\subsection{Matt \& Toni}	
Matt and Toni \shortcite{MT08} compute the strength of an argument using a  two-person zero-sum strategic game. 
This game confronts two players, a proponent and an opponent of a given argument, where the strategies of the players are sets of arguments. 
For an $AF=\langle A,R \rangle$ and $a \in A$, the sets of strategies for the proponent and opponent are $S_{P}(a)=\{P\ |\ P \subseteq A, a \in P\}$ and $S_{O}=\{O\ |\ O \subseteq A\}$ respectively.

\begin{definition}
	Let $F=\langle A,R \rangle$ be an AF and $X,Y \subseteq A$. The {\bf set of attacks from $X$ to $Y$} is defined by $Y_{F}^{\leftarrow X} = \{(a,b)\in X \times Y\ |\ (a,b)\in R\}$. The {\bf degree of acceptability} of $P$ w.r.t $O$ is given by $\phi(P,O) = \frac{1}{2}[1+f(|O_{F}^{\leftarrow P}|)-f(|P_{F}^{\leftarrow O}|)]$ where $ f(n) = \frac{n}{n+1} $.
\end{definition}

\begin{definition} Let $F=\langle A,R \rangle$ be an AF. The {\bf rewards of $P$}, denoted by $r_{F}(P,O)$, are defined by :
		\[
			r_{F}(P,O) = 
			\left\{
			\begin{array}{l l}
			0 & \text{iff}\ \exists a,b \in P, (a,b) \in R,\\
			1 & \text{iff}\ |P_{F}^{\leftarrow O}| = 0,\\
			\phi(P,O) & \text{otherwise}
			\end{array}
			\right.
		\]
\end{definition}

Proponent and opponent choose mixed strategies, according to some probability distributions, respectively $p = (p_1, p_2, \ldots, p_m)$ and $q = (q_1, q_2, \ldots, q_n)$, with $m = |S_{P}|$ and $n = |S_{O}|$. 
For each argument $a \in A$, the proponent's expected payoff $E(a,p,q)$ is then given by
$E(a,p,q) = \sum^{n}_{j=1}\sum^{m}_{i=1}p_{i}q_{j}r_{i,j}$.
Finally the value of the zero-sum game for an argument $a$, denoted by $s(a)$, is
$s(a) = \max_{p}\min_{q} E(a,p,q)$.

\begin{definition}
The {\bf ranking-based semantics M\&T} associates to any AF = $\langle A,R \rangle$  a ranking $\succeqmt_{AF}$ on $A$ such that $\forall a,b \in A, a \succeqmt_{AF} b$ iff $s(a) \ge s(b)$.
\end{definition}

\addtocounter{ex}{-1}
\begin{ex}[cont.]
		One obtains $s(a)\approx0.17$ , $s(b)=1$ , $s(c)=0.25$, $s(d)=0.25$  and $s(e)=0.5$ and the following  preorder: $b \succmt e \succmt c \simeqmt d \succmt a$.
\end{ex}

On this example, we can see that once again the strength of attackers is more important than their numbers ($e$ is ranked higher than $d$). 

\begin{prop}
	The ranking-based semantics 
	M\&T satisfies Abs, In, VP, +AB, SC, Tot, NaE and AvsFD.
	Other properties are not satisfied.
	\end{prop}

\section{Discussion}

As it can be easily checked on the running example, all these proposed ranking semantics have distinct behaviors (the ranking obtained is different for each semantics - see the summary in Table \ref{summary}): this justifies the need of some axiomatic work.
 
		\renewcommand{\arraystretch}{1.4}
		\begin{table}[!ht]
			
			\begin{center}
			\scalebox{0.9} {
			\begin{tabular}{|c|c@{}c@{}c@{}c@{}c@{}c@{}c@{}c@{}c|}
				\cline{1-10}
				\emph{Semantics} & \multicolumn{9}{c|}{\emph{Order between arguments}} \\
				\cline{1-10}
				Cat & $b$ & $\succcat$ & $d$ & $\succcat$ & $e$ & $\succcat$ & $c$ & $\succcat$ & $a$  \\
				\cline{1-10}
				SAF & $b$ & $\succsaf$ & $e$ & $\succsaf$ & $d$ & $\succsaf$ & $c$ & $\succsaf$ & $a$\\
				\cline{1-10}
				M\&T & $b$ & $\succmt$ & $e$ & $\succmt$ & $c$ & $\simeq^{\mbox{\tiny M\&T}}$ & $d$ & $\succmt$ & $a$ \\
				\cline{1-10}
				Dbs & $b$ & $\succdbs$ & $d$ & $\succdbs$ & $c$ & $\succdbs$ & $e$ & $\succdbs$ & $a$\\
				\cline{1-10}
				Bds & $b$ & $\succbbs$ & $d$ & $\succbbs$ & $c$ & $\succbbs$ & $e$ & $\succbbs$ & $a$\\
				\cline{1-10}
			\end{tabular}
 			}	
 			\end{center}
		\caption{Orders obtained on the Example 1 \label{summary}}
		\end{table}
		
Our work initiates this study, by checking properties that have been proposed in the papers that introduce the different semantics. Our analysis is applied to existing semantics, but any new semantics could be inspected through the same lens.\footnote{For instance, the semantics very recently proposed in \cite{GM15}.}
Table \ref{resumeENTIER} summarizes the properties satisfied by the ranking semantics we consider in this paper. 
We also checked what are the properties satisfied by the usual Dung's Grounded semantics, that gives some hints on the compatibility of these properties with classical semantics. Note that, in this case, this is a degenerate ranking semantics with only two levels (accepted/rejected): 

\begin{prop}
The grounded semantics satisfies Abs, In, CT, QP, Tot, NaE, AvsFD. 
Other properties are not satisfied.
\end{prop}
A cross $\times$ means that the property is not satisfied, symbol $\checkmark$ means that the property is satisfied, symbol $−-$ means that the property can not be applied to the semantics (because the semantics is not compatible with the constraint given by the rule), and the shaded cells highlight the results already proved in the literature.
\renewcommand{\arraystretch}{1.1}
\begin{table}[!ht]
		\begin{center} 
		\scalebox{0.77} {
			\begin{tabular}{|c|c|c|c|c|c|c|c|}
				\hline
				Properties & SAF & Cat & Dbs & Bbs & $\text{Tuples}^{*}$ & M\&T & Grounded\\
				\hline \hline
				Abs & \vrai &  \cellcolor{gray!50}\vrai & \cellcolor{gray!50}\vrai & \cellcolor{gray!50}\vrai & \vrai & \vrai  & \vrai\\
				\hline
				In & \vrai &  \cellcolor{gray!50}\vrai & \cellcolor{gray!50}\vrai & \cellcolor{gray!50}\vrai & \vrai & \cellcolor{gray!50}\vrai & \vrai \\
				\hline
				VP & \vrai &  \cellcolor{gray!50}\vrai & \cellcolor{gray!50}\vrai & \cellcolor{gray!50}\vrai & \cellcolor{gray!50}\vrai  & \cellcolor{gray!50}\vrai & \faux\\
				\hline
				DP & \vrai &  \cellcolor{gray!50}\vrai & \cellcolor{gray!50}\vrai  & \cellcolor{gray!50}\vrai & \faux & \faux  & \faux \\
				\hline
				CT & \vrai &  \cellcolor{gray!50}\vrai & \cellcolor{gray!50}\vrai & \cellcolor{gray!50}\vrai & \faux & \faux & \vrai \\
				\hline 
				SCT & \vrai &  \cellcolor{gray!50}\vrai & \cellcolor{gray!50}\vrai & \cellcolor{gray!50}\vrai & \faux & \faux & \faux\\
				\hline
				CP & \faux & \cellcolor{gray!50}\faux & \cellcolor{gray!50}\vrai & \cellcolor{gray!50}\vrai & \faux & \faux  & \faux\\
				\hline
				QP & \faux & \cellcolor{gray!50}\faux & \cellcolor{gray!50}\faux & \cellcolor{gray!50}\faux & \faux & \faux  & \vrai \\
				\hline
				DDP & \faux & \cellcolor{gray!50}\faux & \cellcolor{gray!50}\faux & \cellcolor{gray!50}\vrai & \faux & \faux  & \faux \\
				\hline
				\hline
				SC & \faux & \faux & \faux & \faux & - & \cellcolor{gray!50}\vrai & \faux  \\
				\hline
				\hline
								$\oplus$DB & \faux & \faux & \faux & \faux & \faux & \faux & \faux  \\
								\hline
				+DB & \faux & \faux & \faux & \faux & \cellcolor{gray!50}\vrai & \faux  & \faux \\
				
				\hline
				$\uparrow$AB & \vrai & \vrai & \vrai & \vrai & \cellcolor{gray!50}\vrai & \faux & \faux  \\
				\hline
				$\uparrow$DB & \vrai & \vrai & \vrai & \vrai & \cellcolor{gray!50}\vrai & \faux  & \faux\\
				\hline
				+AB & \vrai & \vrai & \vrai & \vrai & \cellcolor{gray!50}\vrai & \vrai & \faux \\
				\hline
				\hline
				Tot & \vrai & \vrai & \vrai & \vrai & \cellcolor{gray!50}\faux & \vrai & \vrai\\
				\hline
				NaE & \vrai & \vrai & \vrai & \vrai & \vrai & \vrai &  \vrai\\
				\hline
				AvsFD & \faux & \faux & \faux & \faux & \vrai & \vrai & \vrai \\
				\hline
			\end{tabular} 
			}
		\caption{\label{resumeENTIER} Properties satisfy by the studied ranking semantics. }
		\end{center}
	\end{table}

There are a number of observations that we can make regarding these axioms and the results reported in Table \ref{resumeENTIER}: 


\noindent $\bullet$ \emph{Some axioms seem to be widely accepted and shared by all semantics}. 
We see that the properties Abs, In and VP are satisfied by all the ranking semantics. This is expected, since these properties really seem necessary for a good ranking semantics. {Indeed, we recall that the input is a Dung's abstract argumentation framework where there is no information about the nature of arguments, so only the attacks have to be taken into account, hence the importance of Abs. Concerning the property Independence (In), no justification could explain the fact that an argument can influence others arguments without an existing link between them. Finally, the non-attacked arguments are obviously the best arguments in an AF, it is why VP is necessary.}\\
NaE is also satisfied by all semantics. This is also a very basic requirement for a ranking semantics, it mainly says that the non-attacked arguments are all equivalent. This is a kind of compatibility principle with usual Dung's semantics, and it says that only your attackers should impact your ranking, not the arguments you attack.%
\footnote{Note that it could make sense to make a distinction between arguments that attack a lot of arguments and the ones that do not --- so to violate NaE, in particular. This could be considered as some kind of power index. But this is not the aim of ranking semantics.}
Another property that we consider as a requirement is the Tot property, which is in line with the idea of ``ranking'' semantics.  It would be necessary if one wants to use these semantics in real applications. 
This is a drawback of Tuples$^{*}$. 
%
An interesting question is to know if it is possible to refine Tuples$^{*}$, i.e. to define a semantics close to Tuples$^{*}$, but that is easily computable for argumentation frameworks with cycles, and that satisfies Tot. 
A last property satisfied by all semantics is +AB, which states that adding an attack branch towards an argument degrades its ranking. 
This also seems to be a perfectly natural requirement for ranking semantics: the more you are attacked, the worse you are.

Overall, this gives us a set of 6 properties that should be satisfied by any ranking semantics: Abs, In, VP, NaE, Tot and +AB.
One can note that Abs, In, NaE and Tot are satisfied by the grounded semantics, so they are compatible with usual Dung's semantics.
VP and +AB are not satisfied by the grounded semantics, because it only has two levels of evaluation (accepted/rejected), and these two properties really introduce graduality in the evaluation.

\begin{figure}[!ht]
	\begin{center}
		\begin{tikzpicture}[->,>=stealth,shorten >=1pt,auto,node distance=1.1cm,thick,
									main node/.style={circle,draw,font=\small,inner sep=2.3pt},
									sec node/.style={circle,draw,font=\small,inner sep=3pt},
									third node/.style={circle,draw,font=\small,inner sep=1.3pt},
									]

  		 \node[main node] (1) {$a_{4}$};
  		 \node[main node] (2) [right of=1] {$a_{3}$};
 		 \node[sec node] (3) [right of=2] {$a$};
 		 \node[main node] (4) [above of=1] {$a_{2}$};
 		 \node[main node] (5) [right of=4] {$a_{1}$};
 		 \node[main node] (6) [below of=1] {$a_{6}$};
  	     \node[main node] (7) [right of=6] {$a_{5}$};
  	     \node[main node] (8) [below of=6] {$a_{8}$};
  	     \node[main node] (9) [right of=8] {$a_{7}$};
  	     \node[main node] (10) [right of=3] {$b_{1}$};
  	     \node[sec node] (11) [right of=10] {$b$};
  
  		  \draw[->,>=latex] (1) to (2);
  		  \draw[->,>=latex] (2) to (3);
		  \draw[->,>=latex] (4) to (5);
  		  \draw[->,>=latex] (5) to (3);
  		  \draw[->,>=latex] (6) to (7);
  		  \draw[->,>=latex] (7) to (3);
  		  \draw[->,>=latex] (8) to (9);
  		  \draw[->,>=latex] (9) to (3);
  		  \draw[->,>=latex] (10) to (11);
		\end{tikzpicture} 	
		\caption{\label{figureAvsFD}AF that illustrates the property AvsFD}
	\end{center}
\end{figure}
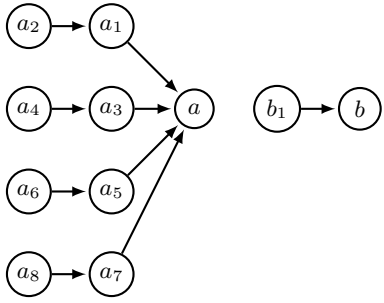

\noindent $\bullet$ \emph{Some axioms are very discriminatory and provide a rough classification of semantics.}
As a general comment, one can check in the table that SAF, Cat, Dbs and Bds share a lot of properties. Tuples$^{*}$ and M\&T seem to belong to another class of semantics: they are the only ones that satisfy AvsFD.
The property AvsFD, illustrated in Figure \ref{figureAvsFD}, which states that an argument that is (only) attacked once by a non-attacked argument (it is the case of $b$ only attacked by $b_{1}$) is worse than an argument that have any number of attacks that all belong to defense branches (it is the case of $a$ which have four defense branches and no attack branch), is a very discriminating property. 
So this property can be seen as a kind of boundary between two sub-classes of ranking properties.
The ones that satisfy it take care of the whole branches of attack/defense. Whereas for the properties that do not satisfy it, a defense branch (that still ends by an attack towards the argument) always penalizes it. 

\noindent $\bullet$ \emph{More specific properties.}
As mentioned already, axioms operate at different levels. We observe that 
 `local' axioms (CP, QP, DP, (S)CT), just looking at direct attackers (or defenders), make choice which can be justified in some situations, but which seem hardly general (and sometimes impossible to reconcile with some more global properties, as our Prop. \ref{prop:incomp} shows). 
Properties related to `change' ($\oplus$DB, +DB, $\uparrow$AB, $\uparrow$DB, +AB) are very appealing. One of our contribution is to have systematically generalized them.

\noindent $\bullet$ \emph{Defining axiomatically the worst arguments is not obvious.}
Interestingly, while all semantics agree axiomatically on which arguments should be the best in a system (VP), there is no consensus regarding the \emph{worst} arguments. SC is very interesting in that respect. 
It makes the observation that a self-contradicting argument is intrinsically flawed, without even requiring other arguments to defeat it. 
But as can be observed none of the semantics comply with it, except the one of Matt and Toni. It is because all semantics consider that an argument that attacks itself is a path like the other ones. So an argument which attacks itself (and by no other argument) is better than an argument which is attacked several times. 

\noindent $\bullet$ \emph{The interplay of axioms is often instructive.} We have identified a number of incompatibilities between axioms. 
There is an additional remark that we can make in that respect, that is related to the incompatibility between VP and 
$\oplus$DB. One can easily remark that $\oplus$DB is more general than +DB, and in a sense more natural: the property is stated for {\sl any} cases, it does not treat some arguments (the non-attacked arguments here) differently. But it contradicts VP in this case. +DB is a less ``systematic'' property (it was the original one proposed in \cite{CLS05}) but is compatible with VP : if one accepts that non-attacked arguments should be the best (VP), it cannot be the case that adding a defense branch \emph{always} improve the situation of a given argument.





\noindent $\bullet$ \emph{This set of axioms is yet to be augmented.}
This can be observed by the fact that SAF and Cat satisfy the same set of properties, whereas they have quite different definitions and behavior. This mean that at least one property is lacking in order to discriminate these two operators.


\section{Conclusion}

In this work we proposed a comparative study of existing ranking-based semantics. It turns out that the existing ranking-based semantics exhibit quite different behaviors and satisfy different properties.
We propose to take as basic properties for ranking-based semantics 
Abs, In, VP, NaE, Tot and +AB. We also put forward AvsFD that discriminates two subclasses of semantics.

There is still work needed on the topic. First to propose other ranking-based semantics.  But it is also important to find other logical properties, and to try to characterize classes of semantics with respect to these properties. 
An ambitious research agenda would be to identify situations where controversial axioms are justified or not. 
	
\section{Acknowledgements}
This work benefited from the support of the project AMANDE ANR-13-BS02-0004 of the French National Research Agency (ANR).

	\bibliographystyle{aaai}
	\bibliography{bibliographie}

\end{document}